\documentclass[11pt,a4paper]{article}
\usepackage[hyperref]{acl2020}
\usepackage{times}
\usepackage{latexsym}

\usepackage{microtype}
\usepackage{multirow}
\usepackage{color}

\usepackage{graphicx}
\usepackage{amsmath}
\usepackage{amsfonts,amssymb}
\usepackage{bm}
\usepackage{booktabs}
\usepackage{algorithm}
\usepackage{algorithmicx}
\usepackage{algpseudocode}
\usepackage{dsfont}
\usepackage{subcaption}
\usepackage{xspace}

\title{\HGNN: Modelling Long-distance Node Relations for \\Improving General Graph Neural Networks}
\newcommand\HGNN{HighwayGraph\xspace}

\author{Deli Chen,\textsuperscript{1}
Xiaoqian Liu,\textsuperscript{1}
Yankai Lin,\textsuperscript{2}
Peng Li,\textsuperscript{2}
Jie Zhou,\textsuperscript{2}
Qi Su,\textsuperscript{1}
Xu Sun\textsuperscript{1}
\\ \\
\textsuperscript{1}{MOE Key Lab of Computational Linguistics, School of EECS, Peking University}\\
\textsuperscript{2}{Pattern Recognition Center, WeChat AI, Tencent Inc, China}\\
\{chendeli,liuxiaoqian,sukia,xusun\}@pku.edu.cn,\\
\{yankailin,patrickpli,withtomzhou\}@tencent.com
}

\date{}

\begin{document}
\maketitle
\begin{abstract}

Graph Neural Networks (GNNs) are efficient approaches to process graph structured data. Modelling long-distance node relations is essential for GNN training and applications.
However, conventional GNNs suffer from bad performance in modelling long-distance node relations due to limited-layer information propagation.
Existing studies focus on building deep GNN architectures, which face the over-smoothing issue and cannot model node relations in particularly long distance. 
To address this issue, we propose to model long-distance node relations by simply relying on shallow GNN architectures with two solutions: (1) Implicitly modelling by learning to predict node pair relations (2) Explicitly modelling by adding edges between nodes that potentially have the same label.
To combine our two solutions, we propose a model-agnostic training framework named \HGNN, which overcomes the challenge of insufficient labeled nodes by sampling node pairs from the training set and adopting the self-training method.
Extensive experimental results show that our \HGNN achieves consistent and significant improvements over four representative GNNs on three benchmark datasets. 
\end{abstract}

\section{Introduction}
Graph structured data has proliferated rapidly in real-world applications among various fields, such as social computing~\citep{model_sage,dataset_amazon}, recommendation system~\citep{recomm1,recomm2}, 3D geometric learning~\citep{3dshape1,3dshape2} and biomedical science~\citep{dataset_ppi,dataset_qm9}. Many of these applications can be reduced to the fundamental semi-supervised node classification problem~\citep{semi-supervised}, which aims to predict node labels with limited annotated nodes.

\begin{figure}[t]
\centering
\includegraphics[width=.95\columnwidth]{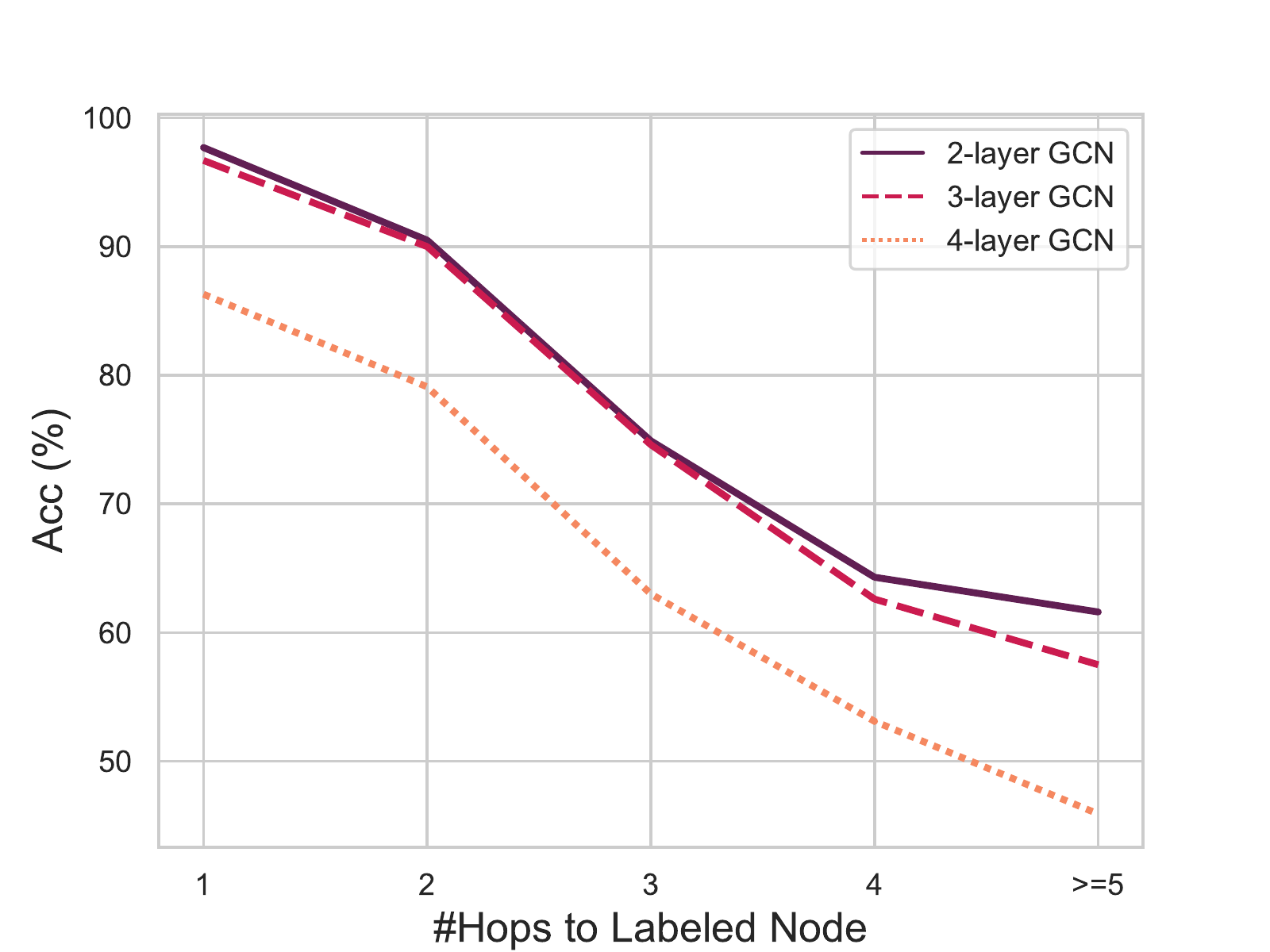}
\caption{Accuracy (Acc) values of test nodes sets with different topological distances to the labeled nodes of the same category for GCN models~\citep{model_gcn} with various layer numbers. \#Hops means the least required hops number for an unlabeled node to contact a labeled node with the same category. It is obvious that the accuracy decreases with the increase of topological distance. Simply adding GNN layers can not improve the model performance for the remote nodes far away from the labeled node.}
\label{figure_intro}
\end{figure}

\begin{figure*}[t]
\centering
\begin{minipage}{0.325\textwidth}
  \centering
  \label{fig:typical}
  \includegraphics[width=\linewidth]{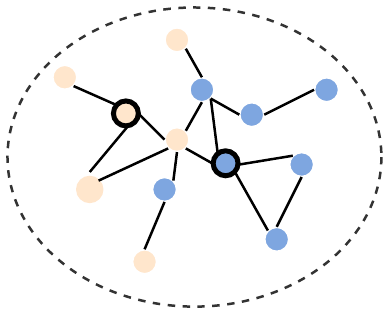}
  \subcaption{Original Graph}
\end{minipage}%
\begin{minipage}{0.325\textwidth}
  \centering
  \label{fig:implicit}
  \includegraphics[width=\linewidth]{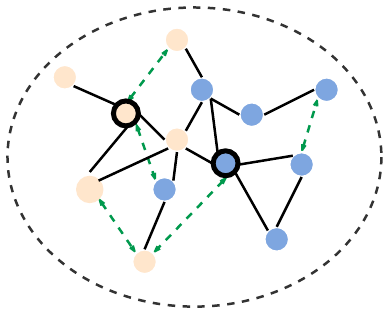}
  \subcaption{Implicit Modelling}
\end{minipage}%
\begin{minipage}{0.325\textwidth}
  \centering
  \label{fig:explicit}
  \includegraphics[width=\linewidth]{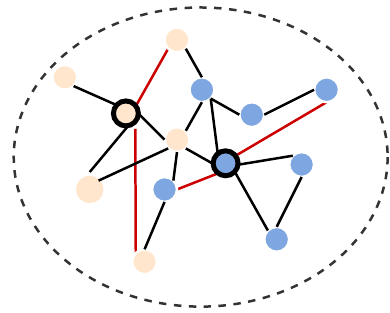}
  \subcaption{Explicit Modelling}
\end{minipage}%
\caption{The schematic diagram for our solutions. Figure (a) displays a virtual graph with nodes in two categories (blue and yellow). Nodes with black border represent labeled nodes. From this figure, we can observe that some nodes are topologically far away from the labeled node, and these nodes are easy to be wrongly predicted according to Figure~\ref{figure_intro}. Figure (b) shows our implicit solution to modelling the long-distance node relation, which implicitly models two nodes by learning their relation (the green dotted lines with arrows means predicting the node pair relation that whether two nodes are of the same category). Figure (c) shows our explicit solution, which directly adds edges between nodes of the same category (the red straight lines represent the added edges for reducing the distance to labeled node).}
\label{figure_idea}
\end{figure*}

Graph neural networks (GNNs) have achieved promising results on semi-supervised node classification by learning node representations through information propagation along edges between nodes~\citep{model_gcn,model_sage,model_hyper_graph, model_feast,model_ggnn,model_stand_graph,model_gmnn,model_hgcn,model_pretrain,model_graphmix}. 
However, a typical GNN architecture has finite hops to propagate information along edges, which limits its capability of modelling the relation between nodes with long topological distance. 
% In this work, we formulate the problem of modelling long-distance node relations as whether labeled nodes (nodes with annotated labels in the training set) and unlabeled nodes (nodes without annotated labels in the testing set) in long topological distance are of the same category. 
We depict the problem in Figure~\ref{figure_intro} using GCN~\citep{model_gcn} as an example. It shows that regardless of the layer number of GCN, the node classification accuracy of the three models all drops substantially as the topological distance from labeled nodes to unlabeled nodes increases (here the labeled node and the unlabeled node are of the same category; similarly hereinafter). Thus, modelling long-distance node relations is an essential and challenging problem for node representation learning and downstream tasks. 

Current studies that aim to overcome limited hops in GNNs for information propagation focus on designing a deep GNN architecture~\citep{deepgcns,drop_edge,PairNorm}. Yet, deep GNNs face the over-smoothing issue~\citep{analysis_smoothing,chen_smoothing} that all node representations will become indistinguishable. More importantly, deep GNNs cannot fundamentally address the problem of modelling long-distance node relations, because the number of hops is still limited for information propagation due to the finite layer number of a deep GNN. Thus, it is infeasible for deep GNNs to model the relation between nodes in a particularly long topological distance.

To solve the problem of modelling long-distance node relations, we propose a new perspective by simply relying on shallow GNNs (i.e., a two-layer GNN). Specifically, we provide two solutions: (1) \textbf{Implicitly modelling} node relations by predicting whether two nodes are of the same category; (2) \textbf{Explicitly modelling} node relations by adding edges between two nodes that potentially have the same category label. 
An illustration of our two solutions is shown in Figure~\ref{figure_idea}. 
Figure~\ref{figure_idea}~(a) shows an original graph consisting of two node categories, and we can observe that some nodes are topologically far away from the labeled nodes, causing a high probability of incorrect prediction according to Figure~\ref{figure_intro}. 
In Figure~\ref{figure_idea}~(b), we implicitly model long-distance node relations by predicting whether two nodes are of the same category. The motivation for this solution is to pass the category information from other nodes to the remote nodes (remote nodes refer to the unlabeled nodes that have a long topological distance to the labeled node) by distinguishing whether they are of the same category.
Besides, Figure~\ref{figure_idea}~(c) shows the explicitly modelling solution, which directly builds the information channel between the remote unlabeled node and the corresponding labeled node by adding edges.

Both of the two solutions are able to model node relations regardless of the topological distance. Thus, the two solutions  enable a GNN to learn better node representations, especially for remote nodes.
However, there is a key challenge in the implementation of the two solutions: there is no label provided for nodes except for the training set in the semi-supervised node classification setting. Thus, it is infeasible to directly conduct the implicit or the explicit modelling. 
To address this issue, we propose a simple yet effective training framework named \textbf{\HGNN} to implicitly and explicitly model long-distance node relations in an indirect way.
For the implicit modelling, we sample node pairs from the training set to estimate long-distance node relations. For the explicit modelling, we adopt the self-training approach by adding edges based on prediction results.

Our \HGNN is model-agnostic and can be applied to any variant of GNNs. Extensive experimental results show that our \HGNN achieves consistent and significant improvements over four representative GNN models on three benchmark graph datasets (CORA, CiteSeer, PubMed) with limited extra computational cost. We also achieve state-of-the-art performance on CORA and CiteSeer.
The main contributions of this work are as follows:
\begin{itemize}
    \item For modelling long-distance node relations in graph structured data, we provide two solutions which simply rely on shallow GNNs: (1) Implicit modelling long-distance node relations by predicting the node pair relation, and (2) Explicit modelling long-distance node relations by adding edges between nodes that potentially have the same label. 
    \item We design a model-agnostic framework named \HGNN based on the implicit and explicit solutions, which overcomes the challenge of insufficient labeled nodes in the semi-supervised learning.
    \item Results show that our \HGNN achieves consistent improvements over four representative GNNs on three benchmark graph datasets and state-of-the-art performance on CORA and CiteSeer, which verifies the effectiveness and the generalizability of our method. 
\end{itemize}

\section{Method}
In this section, we will first introduce tasks for our method, and provide details of our \HGNN based on the implicit and explicit solutions to modelling long-distance node relations.

\subsection{Solution Formalization}
For an undirected graph given the node feature matrix $\bm{X}\in \mathbb{R}^{n \times h}$ and the node adjacency matrix $\bm{A}\in \mathbb{R}^{n \times n}$ ($n$ denotes the node size of the graph and $h$ denotes the dimension size of the initial node embedding), the node classification task aims to train a classifier $f_n$ (usually a GNN model) to distinguish nodes of different categories:
\begin{equation}
\hat{\bm{l}} = f_n(\bm{X},\bm{A}) \label{form1}
\end{equation}
where $\hat{\bm{l}} \in \mathbb{R}^n$ is the predicted category label for all nodes. 
Different from the node classification task, we design a new task named as {\em node pair classification} to assist in training GNNs. The node pair classification task aims to train a different classifier $f_p$ to predict whether two nodes are of the same category:
\begin{equation}
\hat{\bm{r}} = f_p(\bm{X},\bm{A})  \label{form2}
\end{equation}
where $\hat{\bm{r}} \in \mathbb{R}^{n \times n}$ and $\hat{\bm{r}}_{i,j} \in \{0,1\}$ ($0$ means the $i$-the and $j$-th nodes are of different categories and $1$ denotes they are of the same category).
% The target 
Usually, the total number of all node pairs ($n \times n$) is too large to be enumerated, thus we need a sampling strategy $\mathrm{Sample}(\cdot)$ to sample some node pairs for loss calculation. 
For the prediction in Eq~\ref{form1},~\ref{form2}, we calculate the training loss as:
\begin{flalign}
 \mathcal{L}_{node} &= \mathrm{LossFunc_1}(\hat{\bm{l}},\bm{l}) \label{lf1} \\
 \hat{\bm{r}}' &= \mathrm{Sample}(\hat{\bm{r}}) \label{sample} \\
 \mathcal{L}_{pair} &=\mathrm{LossFunc_2}(\hat{\bm{r}}',{\bm{r}'}) \label{lf2}
\end{flalign}
$\mathrm{LossFunc_1}$ and $\mathrm{LossFunc_2}$ are the loss functions for the node and the node pair classification tasks, respectively. $\bm{l}$ is the gold label for the node classification task and $\bm{r}'$ is the gold label of sampled node pairs for the node pair classification task.
According to Figure~\ref{figure_idea}~(b), node pair prediction enables to model relations between any node pairs, which transfers the category information between nodes. Therefore, we propose to predict the node category label and the node pair relation in a single model, and combine $\mathcal{L}_{pair}$ and $\mathcal{L}_{node}$ to form the loss function as: 
\begin{equation}
    \mathcal{L} = \mathcal{L}_{node} + \lambda \cdot \mathcal{L}_{pair} \label{form_loss}
\end{equation}
where $\lambda$ is the parameter to control the influence of the node pair classification task. When $\lambda = 0$, it falls back to conventional GNN training. 

\begin{algorithm}[t]
% \label{algorithm}
\centering
\footnotesize

\begin{algorithmic}[1]
\caption{HighwayGraph}
\label{framework}
\Require A GNN:$\large{g}(\bm{X},\bm{A})$.
Node feature matrix $\bm{X}$. 
Adjacency Matrix $\bm{A}$.
Number of epochs $N$. 
Number of max iteration time $max_t$.
Gold label for node and node pair $\bm{Y,P}$.
Loss functions for node and node pair$\mathcal{L}_1,\mathcal{L}_2$.
Node pair loss weight $\lambda$.
Mask matrix $\mathbf{Mask}$.
% Nodes indexes for train/valid/test set $[train],[vallid],[test]$.
% Pair indexes to select sampled node pairs $[pair]$.
% Parameter $\lambda$ to control the impact of the node pair classification task.
%  are the nodes indexes to select the nodes in  is the node pair indexes to select the sampled node pairs.
% \State $\mathrm{acc_0} \gets \large{g}(X,A)$
\State $\mathrm{valid\_acc_0} = 0 $
\For{iter times $ \mathrm{i} \in [1,max_t]$}
\For{$t \in [1,N]$}
\State $\mathbf{l}$, $\mathbf{r}$ = $\large{g_i}(\bm{X},\bm{A})$ 
\Comment{\textit{Predict node category label and node pair relation jointly}}

\State $\mathcal{L}_{node} = \mathcal{L}_1\big(\mathbf{l}_{[train]}, \bm{Y}_{[train]}\big)$ 
% \Comment{\textit{Calculate node classification loss}}
\State $\mathcal{L}_{pair} = \mathcal{L}_2\big(\mathbf{r}_{[pair]}, \bm{P}_{[pair]}\big)$ 
% \Comment{\textit{Calculate node pair classification loss}}
\State $\mathcal{L}= \mathcal{L}_{node} + \mathcal{\lambda}*\mathcal{L}_{pair}$ 
% \Comment{\textit{Calculate co-training loss}}
\State Back propagate $\mathcal{L}$
\EndFor
\State $\mathrm{valid\_acc_i} \gets \mathrm{Accuracy}(\mathbf{l}_{[valid]},\bm{Y}_{[valid]})$
\If{$\mathrm{valid\_acc_i} \le \mathrm{valid\_acc_{i-1}}$}
\State $\mathrm{test\_acc_i} \gets \mathrm{Accuracy}(\mathbf{l}_{[test]},\bm{Y}_{[test]})$
\State $\mathbf{return\,} \mathrm{test\_acc_{i}} $
\EndIf
\State $\bm{R}^n \gets \mathbf{l\times l}$ 
% \Comment{\textit{Obtain node classification prediction matrix}}
\State $\bm{R}^p \gets \mathbf{r}$ 
% \Comment{\textit{Obtain node pair classification prediction matrix}}
\State $\bm{R}  \gets \bm{R}^n \wedge \bm{R}^p$ 
\Comment{\textit{Obtain prediction matrix for adding edges by the joint decision of node and node pair prediction.}}
\State $\bm{R}' \gets \bm{R} \circ \mathbf{Mask}$ 
\Comment{\textit{Select position for edge addition}}
\State $\bm{A} \gets \bm{A}   \vee \bm{R}'$ 
\EndFor
\State $\mathrm{test\_acc_i} \gets \mathrm{Accuracy}(\mathbf{l}_{[test]},\bm{Y}_{[test]})$
\State $\mathbf{return\,} \mathrm{test\_acc_{i}} $
% \Comment{\textit{Add edges between two nodes of the same class}}
% \State $\mathrm{acc_i} \gets \large{g}(X,\mathbf{A'_{i-1}})$
% \If{$\mathrm{acc_i} \le \mathrm{acc_{i-1}}$}
% \State $\mathbf{return\,} \mathrm{acc_{i-1}} $
% \EndIf
% \EndFor
% \State $\mathbf{reutrn\,} \mathrm{acc_{max_t-1}} $
\end{algorithmic}
\end{algorithm}

\subsection{\HGNN}
To combine the implicit solution and the explicit solution, we propose \HGNN for training GNNs. The full algorithm of \HGNN is presented in Algorithm~\ref{framework}.
A key challenge for implementing the implicit and the explicit solution is that there is no label provided for nodes except for the training set in the semi-supervised learning.
Thus, \HGNN is a self-training approach with multiple iterations.
During one training iteration, \HGNN learns the node and node pair classification jointly (the node pairs are extracted from the training set). 
Then we optimize the original graph topology by adding edges between intra-category nodes according to the joint decision of node and node pair prediction results. 
We then train the next iteration based on the optimized graph topology.
The following sections will elaborate the implicit and explicit ways of modelling long-distance node relations in \HGNN.

\subsubsection{Implicit Modelling: Co-training with Node Pair Classfication}
% In our \HGNN, the implicit way to model long-distance node relations is to co-train the node and node pair classification tasks. We introduce node pair classification as an auxiliary task for two purposes: (1) to strengthen intra-category relations; (2) to weaken inter-category relations. Formally, 
Given the node embedding $\bm{X}$, the adjacency matrix $\bm{A}$ (either an original graph or an optimized graph), and a GNN model ${\rm GNN_{\alpha}}$, we first get hidden representations of nodes:
\begin{flalign}
 \bm{y} &= {\rm GNN_{\alpha}}(\bm{X},\bm{A}) \label{y1} 
\end{flalign}
Then we predict the node category label and the node pair relation based on $\bm{y}$: %node representations $\bm{y}$:
\begin{flalign}
 \hat{\bm{l}} &= {\rm softmax}(\bm{y}) \label{re1} \\
 \hat{\bm{r}} &= {\rm sigmoid }(\bm{y} \cdot \bm{y}^\top)  \label{re2}
\end{flalign}
where $\hat{\bm{l}} \in \mathbb{R}^{n}$ is the predicted node category label and $\hat{\bm{r}} \in \mathbb{R}^{n \times n}$ is the predicted node pair relation. $n$ denotes the node size of the graph. $\hat{\bm{r}}_{i,j} \in [0,1]$ represents the relation between the $i$-th node and the $j$-th node: when $\hat{\bm{r}}_{i,j}$ is close to 1, two nodes are more likely to be of the same category, and vice versa.

The sampling strategy (introduced in Eq~\ref{sample}) used in \HGNN is sampling all the node pairs in the training set, which are the only available labeled nodes during the training. 
The size of the training set is usually limited in the semi-supervised learning, thus we keep all the node pairs in the training set.
As introduced in Eq~\ref{lf1},~\ref{lf2}, we calculate the Negative Log Loss ($\mathrm{LossFunc_1}$) and the Binary Cross Entropy Loss ($\mathrm{LossFunc_2}$) for node label and node pair relation predictions, respectively:
\begin{flalign}
\mathcal{L}_{node} = &-\sum^T_{i=1} \bm{l}_i\,\log\,p(\,\hat{\bm{l}}_i) \label{l1} \\
\begin{split}
 \mathcal{L}_{pair} = &-\sum^T_{i=1}\sum^T_{j=1}\{ \bm{r}_{i,j}\cdot {\rm log} (\hat{\bm{r}}_{i,j}) \\
& + (1-\bm{r}_{i,j}) \cdot {\rm log} (1- \hat{\bm{r}}_{i,j})\} \label{l2}
\end{split}
\end{flalign}
where $T$ is the size of the training set; $\bm{l}$ and $\bm{r}$ are the gold labels for the node category and the node pair relation, respectively; $\bm{r}$ is extracted from $\bm{l}$ by identifying whether two nodes are of the same category. The final training loss is defined in Eq.~\ref{form_loss}.

In experiments, we also use different GNN architectures to co-train the two tasks, such as leveraging two different GNN layers after sharing the same first layer, or adding a trainable parameter matrix in Eq~\ref{re2}. However, these techniques lead to performance degradation and extra training cost. 

\subsubsection{Explicit Modelling: Adding Intra-category Edges}
Besides the implicit solution by co-training with node pair classification, we also adopt an explicit way to model long-distance node relations in \HGNN by adding edges between nodes of the same category.
Since most node labels are not available in the semi-supervised training, we use the self-training method and add edges based on the joint decision of node classification and node pair classification prediction results.
For node label prediction in Eq~\ref{re1}, we get the node label prediction matrix $\bm{R}^n$ by considering the predicted labels and the confidence of two nodes:
\begin{equation}
\bm{R}^n_{i,j}=
\begin{cases}
1,& \hat{\bm{l}}_i = \hat{\bm{l}}_j, \bm{c}_i>t_n, \bm{c}_j>t_n\\
0,& {\rm otherwise} \label{r1}
\end{cases}
\end{equation}
where $\bm{c}_i$ is the prediction confidence for the $i$-th node (the max value of tensor after softmax operation in Eq~\ref{re1}). $t_n \in [0,1]$ is the confidence threshold to filter out low-confidence predictions.
For node pair relation prediction in Eq~\ref{re2}, we get the node pair prediction matrix $\bm{R}^p$ calculated with a threshold $t_p\in [0,1]$:
\begin{equation}
\bm{R}^p_{i,j}=
\begin{cases}
1,& \hat{\bm{r}}_{i,j}\geq t_p\\ 
0,& \hat{\bm{r}}_{i,j}< t_p \label{r2}
\end{cases}
\end{equation}
Then we combine $\bm{R}^n$ and $\bm{R}^p$ to make a joint decision and access a more reliable prediction matrix for explicitly modelling long-distance node relations.
We conduct element-wise ${\rm AND\; (\wedge)}$ operation to access the prediction matrix for adding intra-category edges:
\begin{flalign}
\bm{R} & = \bm{R}^n \wedge \bm{R}^p \\
\bm{R}'& = \bm{R} \circ \mathbf{Mask} \\
\bm{A}'& = \bm{A}   \vee \bm{R}'
\end{flalign}
$\vee$ is the element-wise ${\rm OR}$ operation, $\circ$ is the element-wise multiplication operation, $\bm{A}$ is the original adjacency matrix, and $\bm{A}'$ is the updated adjacency matrix used in the next training iteration.
$\mathbf{Mask}$ is a mask matrix to select the position to add edges. 

In the explicit modeling method, we propose to reduce the long topological distance by adding edges between remote nodes and labeled nodes, so we filter out several rows of $\bm{R}$  to add edges by setting the values of these rows in $\mathbf{Mask}$ to be $1$ and all other values  be $0$. In practice, we select one row per category from the training set (i.e. there are 7 categories in CORA dataset, and we select 7 nodes with different labels from the training set and then select the rows corresponding to these 7 nodes from $\bm{R}$). We find that adding too many edges may cause performance decline. One possible reason is that adding too many edges may introduce too much noise (wrongly added edges) which misleads the node representation learning process.

\begin{table*}[]
    \centering
%     \small
    \resizebox{.98\textwidth}{!}{
\begin{tabular}{l|llll|llll|llll}
\toprule
\textbf{Acc (\%)}   & \multicolumn{4}{c|}{\textbf{CORA}}                           & \multicolumn{4}{c|}{\textbf{PubMed}}                       & \multicolumn{4}{c}{\textbf{CiteSeer}}                         \\ \midrule
\textbf{Model}     & \textbf{GCN} & \textbf{GAT} & \textbf{SAGE} & \textbf{Hyper} & \textbf{GCN} & \textbf{GAT} & \textbf{SAGE} & \textbf{Hyper} & \textbf{GCN} & \textbf{GAT} & \textbf{SAGE} & \textbf{Hyper} \\ \midrule
\textbf{Typical GNN Training}   
&   80.1\tiny$\pm$2.0   &    78.5\tiny$\pm$1.0   
&   80.1\tiny$\pm$1.0   &    79.9\tiny$\pm$1.0      
&   76.5\tiny$\pm$1.3   &    75.0\tiny$\pm$1.0      
&   75.3\tiny$\pm$1.6   &    73.6\tiny$\pm$1.5     
&   67.0\tiny$\pm$0.7   &    66.7\tiny$\pm$1.0   
&   66.6\tiny$\pm$1.0   &    64.3\tiny$\pm$0.8\\

% \textbf{AdaEdge} &  80.7\tiny$\pm$0.8          &    79.8\tiny$\pm$1.3          &80.5\tiny$\pm$1.2           &80.3\tiny$\pm$0.9           &67.4\tiny$\pm$1.0           &   67.0\tiny$\pm$1.4           &  67.7\tiny$\pm$0.9             &  \textbf{67.2}\tiny$\pm$0.8      &  76.4\tiny$\pm$1.3       &  75.0\tiny$\pm$2.2            & 75.7\tiny$\pm$1.5       &  74.4\tiny$\pm$1.0          \\ 
\midrule
\textbf{\HGNN(Full Method)} 
&  \textbf{83.2}\tiny$\pm$0.8  &  \textbf{83.4}\tiny$\pm$0.6
&  \textbf{84.2}\tiny$\pm$0.7  &  \textbf{83.0}\tiny$\pm$1.0   
&  \textbf{77.2}\tiny$\pm$0.5   &  \textbf{76.1}\tiny$\pm$1.5
&  \textbf{75.9}\tiny$\pm$1.8   &  \textbf{74.5}\tiny$\pm$1.3
&  \textbf{70.2}\tiny$\pm$0.4   &  \textbf{68.5}\tiny$\pm$0.3
&  \textbf{68.7}\tiny$\pm$0.5    &  \textbf{67.1}\tiny$\pm$0.3

\\
\textbf{\HGNN(\emph{w/o} Joint Deci.)}
&  82.6\tiny$\pm$1.1       &  82.7\tiny$\pm$0.8   
&  83.6\tiny$\pm$0.7       &  82.3\tiny$\pm$1.0      
&  77.1\tiny$\pm$0.6       &  76.1\tiny$\pm$1.6      
&  75.6\tiny$\pm$1.8       &  74.2\tiny$\pm$1.5   
&  69.3\tiny$\pm$0.4       &  68.3\tiny$\pm$0.6       
&  68.2\tiny$\pm$0.7       &  66.0\tiny$\pm$0.9  \\ 

\textbf{\HGNN(\emph{w/o} Explicit Link)} 
&  83.0\tiny$\pm$1.0       &  81.5\tiny$\pm$1.1   
&  82.7\tiny$\pm$1.0       &  81.0\tiny$\pm$0.8       
&  76.9\tiny$\pm$1.3       &  75.6\tiny$\pm$1.2      
&  75.1\tiny$\pm$1.8       &  74.1\tiny$\pm$1.5   
&  68.0\tiny$\pm$0.6       &  67.8\tiny$\pm$0.9       
&  67.8\tiny$\pm$0.9       &  64.9\tiny$\pm$0.9 \\
\bottomrule

\end{tabular}}
\caption{Node classification results (\% test accuracy) compared to the baselines across three benchmark datasets using four representative GNN models with random dataset splitting.
The mean accuracy and the standard deviation of each experiment are calculated after 15 running times (5 random dataset splitting and 3 random initial seeds for each splitting). Experiments in the same column adopt the same GNN model and set the same hyper-parameters for comparison. We can find that our \HGNN brings significant and consistent performance improvements for all the graph datasets and models.}
%same GNN model for each training iteration in our Full Method to verify the effectiveness of our method in different GNN models
%We limit the GNN architecture trained in each iteration to be the same, while better results can be achieved by selecting different GNN architectures in different training iterations.
\label{table_main_result}
\end{table*}

\section{Experiments}
\subsection{Datasets and Models}
We evaluate our proposed \HGNN on three benchmark graph datasets: CORA, CiteSeer and PubMed~\citep{dataset_ccp}. These datasets are citation graph networks, which have been widely used to evaluate GNNs ~\citep{analysis_lowpass, analysis_smoothing,model_arma, model_dna}. %The statistics of these datasets are presented in Table~\ref{table_dataset_intro}.
To verify the generalizability of \HGNN, we select four representative GNN architectures as the experimental models: 
% GCN~\citep{model_gcn}, GAT~\citep{model_gat}, GraphSAGE~\citep{model_sage}, HyperGraph~\citep{model_hyper_graph}. 
\begin{itemize}

\item \textbf{GCN} (Graph Convolutional Network)~\citep{model_gcn} which uses the spectral method to conduct convolution operation.

\item \textbf{GAT} (Graph Attention Network)~\citep{model_gat} which adopts the attention mechanism to aggregate neighborhood node information.% differently.
   
\item \textbf{SAGE} (GraphSAGE)~\citep{model_sage} which uses a sampling method to propagate information in large graphs.

\item \textbf{Hyper} (HyperGraph)~\citep{model_hyper_graph} which utilizes the high-order information of graphs.
\end{itemize}

\subsection{Experimental Settings}
Following~\citet{dataset_amazon,method_fishergcn}, we run $5$ random dataset splittings using the 20/30/rest splitting method\footnote{Each category has $20$ labeled samples for training and $30$ for validation; the rest labeled nodes are used for testing}, and use $3$ random initial seeds for each splitting method in each experiment to reduce the randomness of the results caused by dataset splittings and initial seeds.
The mean accuracy and standard deviation of each experiment for $15$ runs are reported.
To achieve a highly reliable confidence of graph topology optimization, we set the category threshold of the node pair classification to a rather high value (i.e. $0.9$), which aims to achieve a high-precision and low-recall classifier for two nodes of the same category (there is no need to add all the intra-category edges, but the precision of added edges need to be guaranteed). 
The implementation of the GNN models is based on PyTorch~\citep{code_pytorch} and PyTorch Geometric~\citep{code_geometric}, without changing the implementation of the convolutional layer and dataset in PyTorch Geometric.
Hyperparameters of each GNN are tuned in each group (experiments using the same GNN model on the same graph dataset). Then all the hyperparameters (including the splitting seeds and the initial seeds) are fixed in each experiment group for fair comparison. 
For the node pair classification, we increase the weight of positive samples, since the negative samples are far more than the positive samples.

For all the experimental models, we use full-batch training with a maximum number of training epochs $200$. We use early stopping and stop optimization if the validation accuracy is no longer increased in $10$ epochs. We use dropout~\citep{dropout} method to avoid over-fitting and the dropout rate is tuned in  $\{0.3, 0.4, 0.5, 0.6, 0.7\}$ for different combinations of the GNN models and the graph datasets. The hidden layer number of each GNN model is set to $2$, which can achieve the best performance for GNNs. The hidden size of each hidden layer is tuned in $\{32, 64, 128, 256\}$. We use Adam~\citep{adam} as the optimizer; the initial learning rate is selected in $\{5\times10^{-4}, 1\times10^{-3}, 2\times10^{-3}, 3\times10^{-3}\}$.

\subsection{Baselines}
We compare our proposed training framework with the typical GNN training method, as well as two ablation versions of \HGNN
\begin{itemize}

\item \textbf{Typical GNN Training}: GNNs are trained on the original graph topology using standard semi-supervised training framework~\citep{semi-supervised}. 

% \item \textbf{AdaEdge}~\citep{chen_smoothing}: This method simply adds/removes edges based on node classification predictions iteratively. The main differences about the edge operation between AdaEdge and our \HGNN are that (i) AdaEdge adds and/or removes edges for improving the information-to-noise ratio on a random area of the adjacent matrix to alleviate the over-smoothing issue, while \HGNN adds edges between labeled nodes and unlabeled nodes to reduce the topological distance. (ii) AdaEdge conducts edge operations based on only node label predictions, while \HGNN is based on the joint decision of node label and node pair relation predictions.

\item \textbf{\HGNN (\emph{w/o} Joint Decision)}: This ablation version adds edges only based on node label predictions, which is designed to evaluate the effects of co-training with node pair classification.
   
\item \textbf{\HGNN (\emph{w/o} Explicit modelling)}: Another ablation version of \HGNN which only conducts the first-turn training and does not retrain on the updated graph. This baseline is designed to evaluate the effects of both co-training and iterative training operations.
\end{itemize}

\section{Results and Analysis}

\subsection{Overall Results}
The performance of our \HGNN method and other baselines are shown in Table~\ref{table_main_result}. From the results, we can observe that:

(1) Combining both the implicit modelling (co-training with node pair classification) and the explicit modelling (adding edges), our proposed \HGNN especially improves the performance of the typical GNN training method by a large margin among all the four experimental GNNs. The results demonstrate both the effectiveness and the generalizability of \HGNN.

(2) Our \HGNN performs well on the GraphSAGE architecture, which is designed for large-scale graph networks. Thus, our \HGNN method shows great potential for applications involving large graphs.

(3) The \HGNN (\emph{w/o} Explicit Link) method outperforms all the GNNs trained using the typical method, which validates the effect of co-training with node pair classification. 

(4) Removements of the explicit linking and the joint-decision mechanism both cause significant performance degradation, which demonstrates the effectiveness of the two methods in our \HGNN.

\begin{table}[t]
    \centering
%     \small
    \resizebox{.95\columnwidth}{!}{
\begin{tabular}{lc@{}lc@{}lc@{}l}
\toprule
\multicolumn{1}{l}{\textbf{Method}}& \multicolumn{2}{l}{\textbf{CORA}}     & \multicolumn{2}{l}{\textbf{PubMed}} & \multicolumn{2}{l}{\textbf{CiteSeer}} \\ \midrule
GCN*~\citep{model_gcn}   & 81.5   &   & 79.0    &  & 70.3    \\
GAT*~\citep{model_gat}   & 83.0   &   & 79.0   &  & 72.5\\
GMNN*~\citep{model_gmnn}  & 83.7   &   & \textbf{81.8}    &  & 73.1   \\
Trun. Krylov*~\citep{model_krylov} & 83.8   &   & 80.1      &  & 74.2 \\
GraphMIX*~\citep{model_graphmix}& 83.9   &   & 81.0    &  &  74.5  \\
G-APPNP*~\citep{model_pretrain} & 84.3   &   & 81.0     &  & 72.0  \\
Graph U-net*~\citep{unet}   & 84.4   &   &   79.6     &  & 73.2  &   \\
H-GCN*~\citep{model_hgcn}  & 84.5   & &  79.8 &  &  72.8      &   \\ \midrule
\HGNN (Ours) & \textbf{85.7}&\tiny$\pm$ 0.5    &80.1 & \tiny$\pm$0.4&  \textbf{75.1} &\tiny$\pm$0.2  \\
\bottomrule
\end{tabular}}
\caption{Node classification results in comparison with recent state-of-the-art methods with standard dataset splitting~\cite{semi-supervised}. We run 10 turns for each experiment and report the mean value and the standard deviation over the running times. [*] means the results are taken from the corresponding papers. Our \HGNN set the new SOTA for the CORA and CiteSeer datasets.}

\label{table_state-of-the-art}
\end{table}

\begin{figure*}[t]
\centering
\begin{minipage}{0.31\textwidth}
  \centering
  \includegraphics[width=\linewidth]{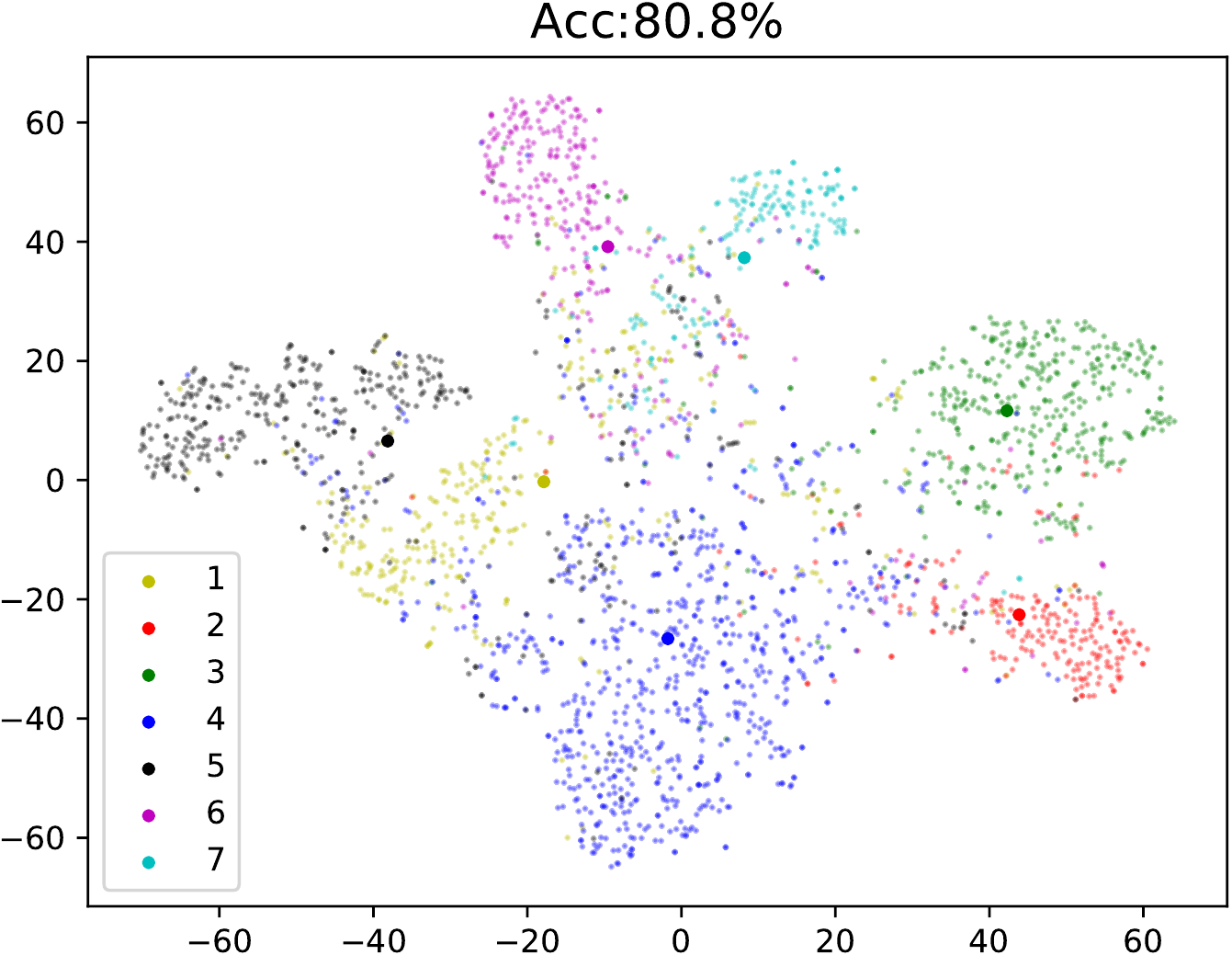}
  %\vspace{-0.05in}
  \subcaption{GCN, $\lambda$ = 0}
\end{minipage}%
 \hspace{0.1in}
\begin{minipage}{0.31\textwidth}
  \centering
  \includegraphics[width=\linewidth]{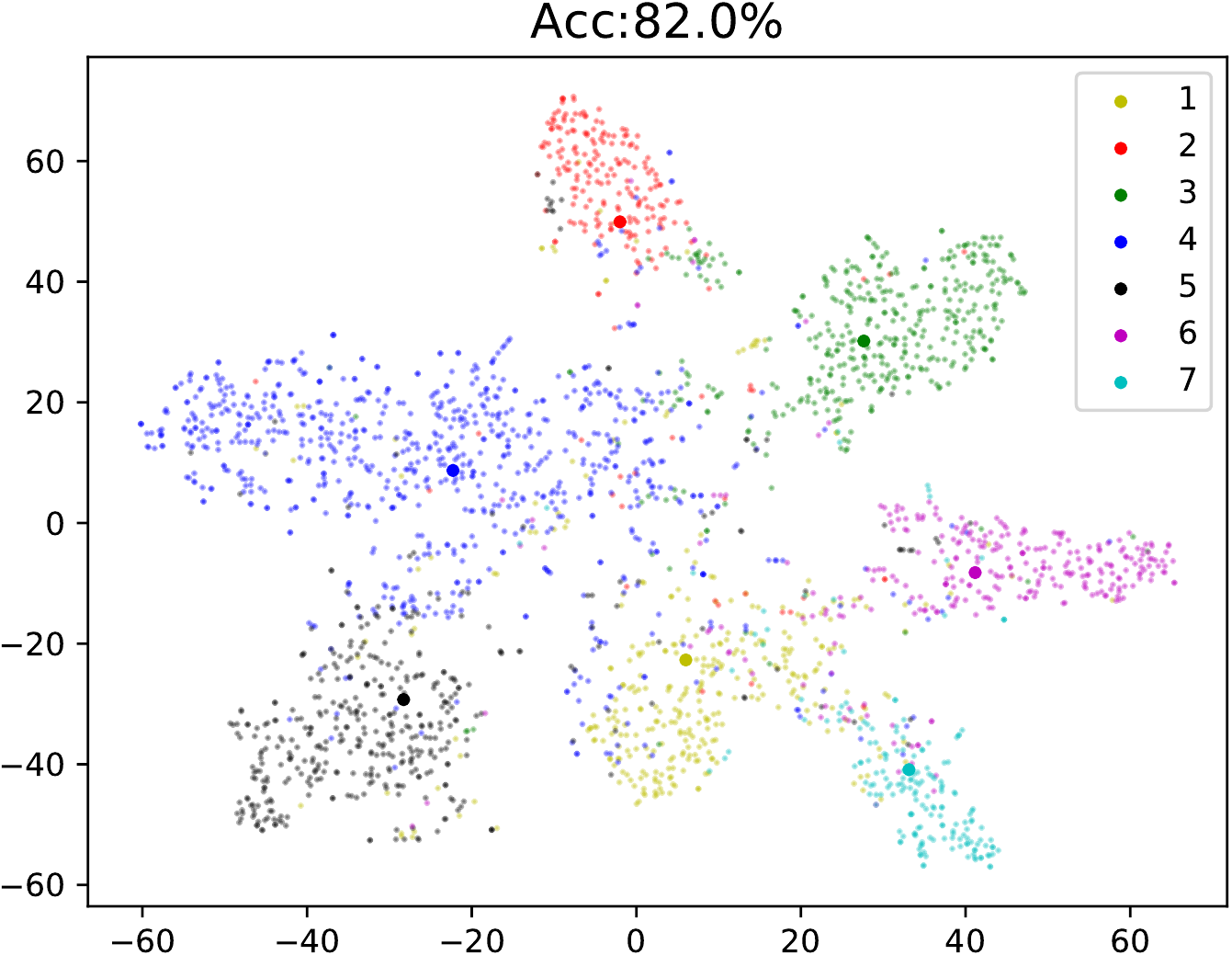}
  %\vspace{-0.05in}
  \subcaption{GCN, $\lambda$ = 0.5}
\end{minipage}%
 \hspace{0.1in}
\begin{minipage}{0.31\textwidth}
  \centering
  \includegraphics[width=\linewidth]{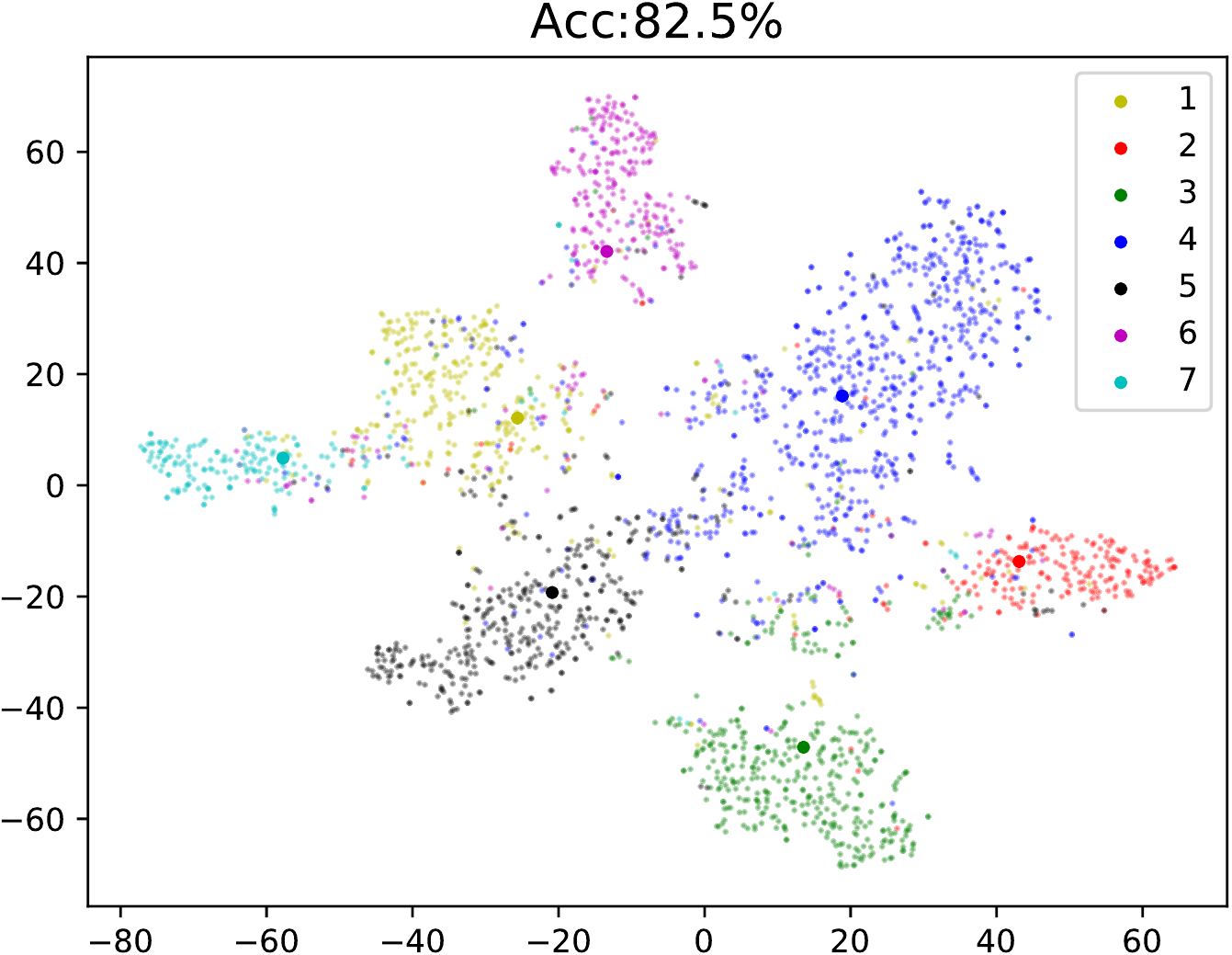}
  %\vspace{-0.05in}
  \subcaption{GCN, $\lambda$ = $1.0$}
\end{minipage}%
%\vspace{0.1in}
\caption{2D visualization (t-SNE) of the nodes distribution in CORA graph. Nodes of the same category share the same color. 
We display the node representations of the second layer in the 2-layer GCN model using our \HGNN framework under different loss weight $\lambda$ of node pair prediction from Eq~\ref{form_loss}.
We can easily find that with the increase of $\lambda$, nodes of the same category are more concentrated and nodes of different classes are more separated. The prediction accuracy is also increased.}
\label{figure_visual}
\end{figure*}

\begin{figure*}[t]
\centering
\begin{minipage}{0.245\textwidth}
  \centering
  \includegraphics[width=\linewidth]{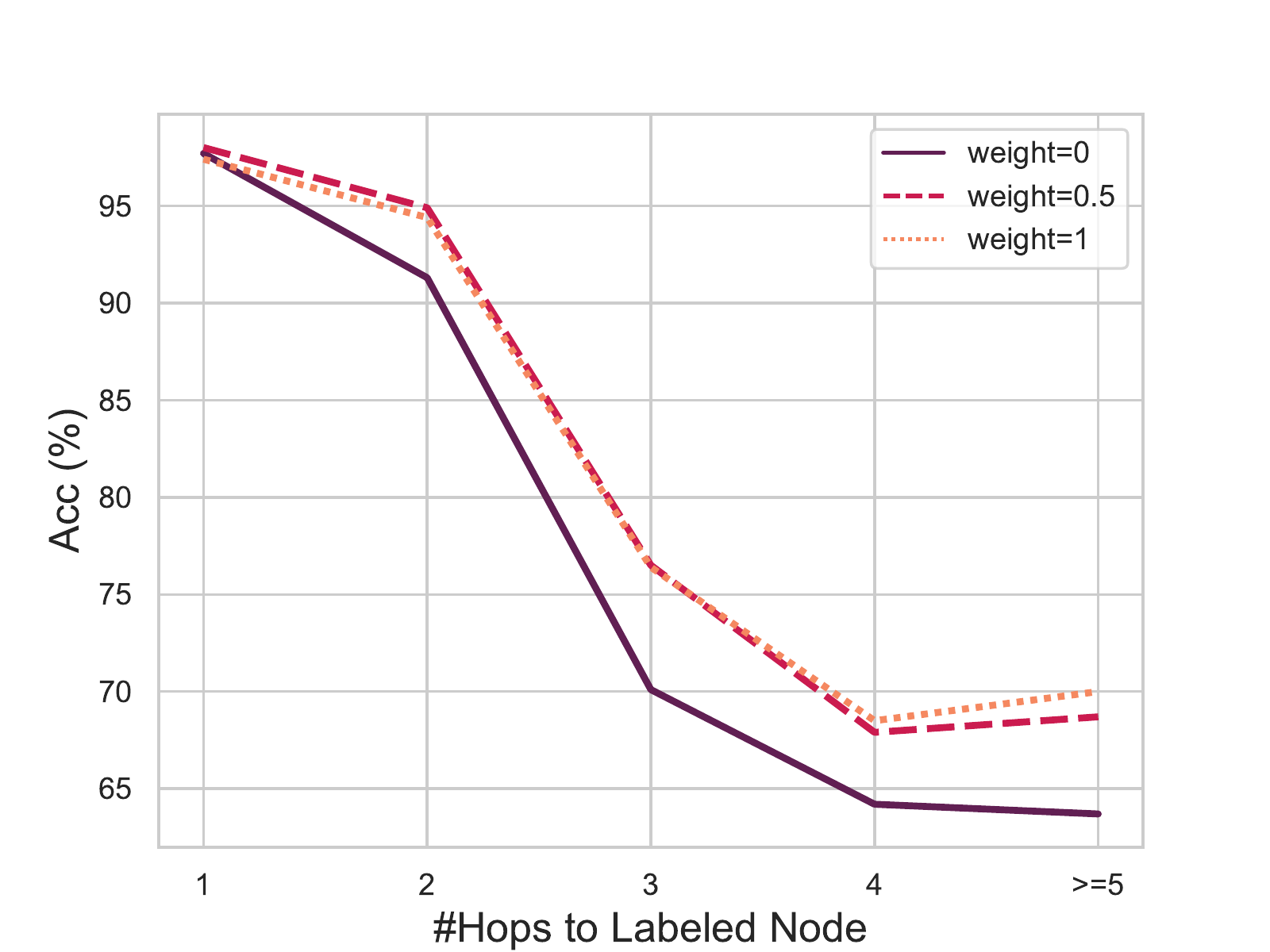}
  \subcaption{GCN}
\end{minipage}%
\begin{minipage}{0.245\textwidth}
  \centering
  \includegraphics[width=\linewidth]{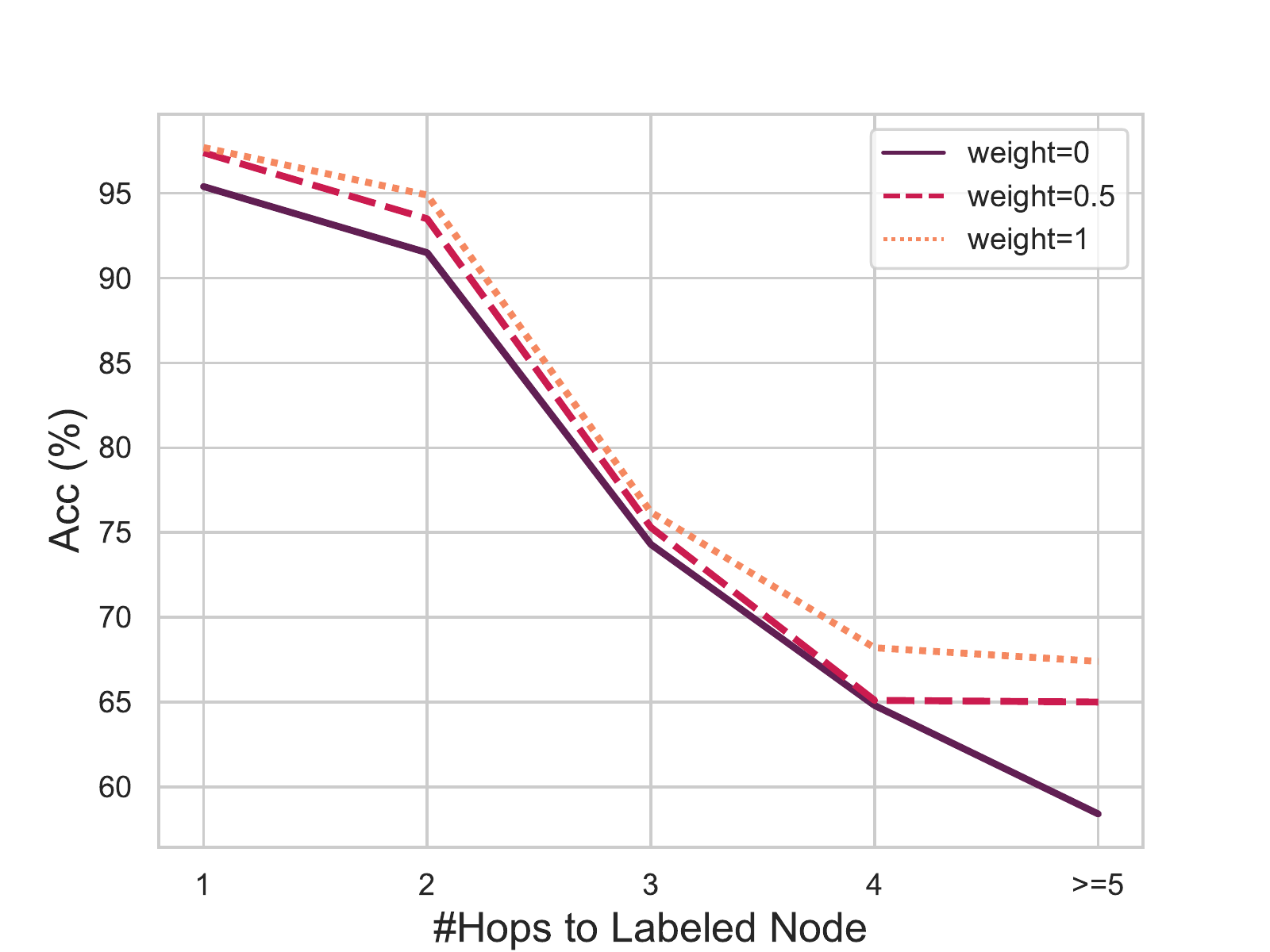}
  \subcaption{GAT}
\end{minipage}%
\begin{minipage}{0.245\textwidth}
  \centering
  \includegraphics[width=\linewidth]{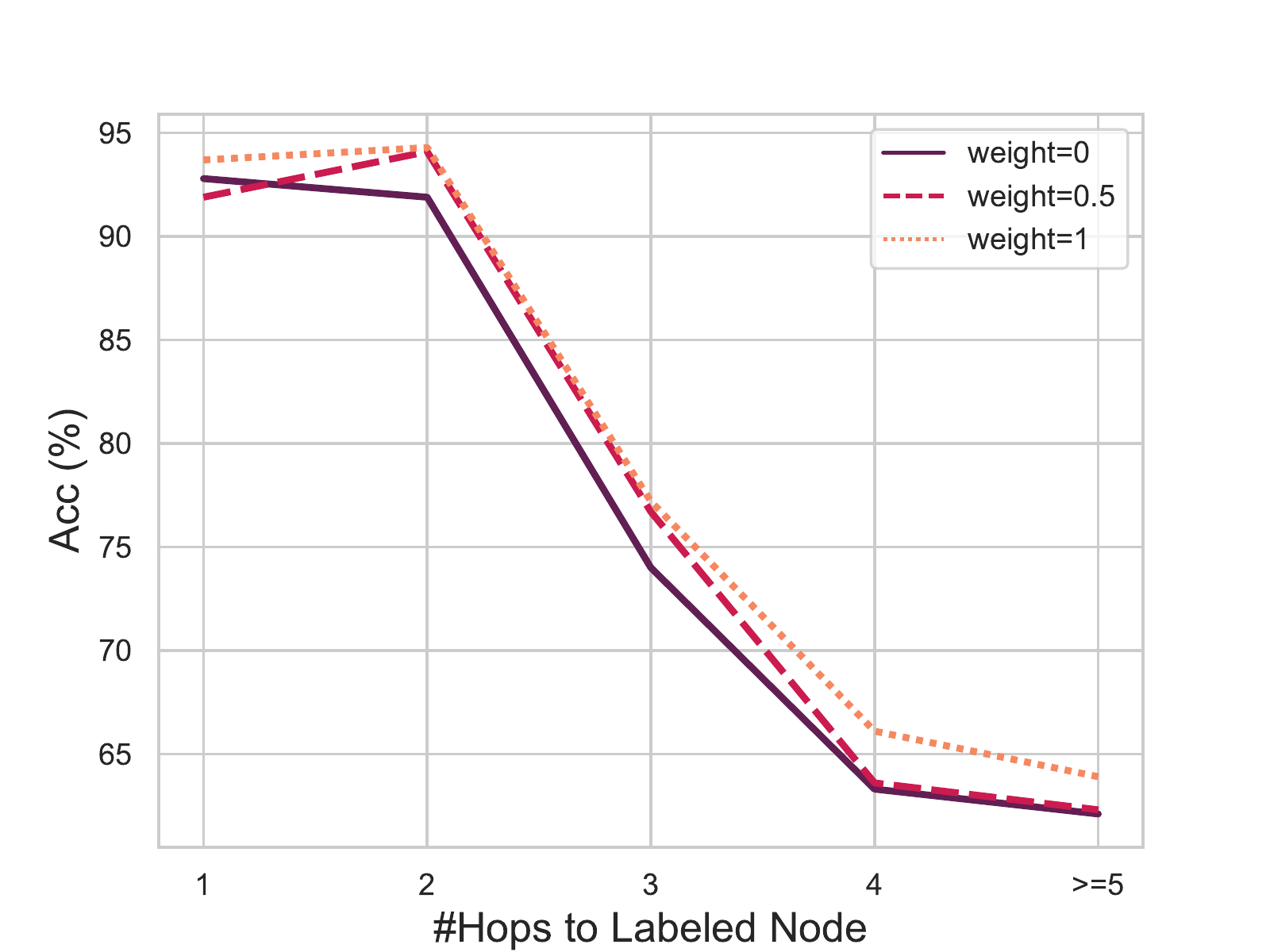}
  \subcaption{HYPER}
\end{minipage}
\begin{minipage}{0.245\textwidth}
  \centering
  \includegraphics[width=\linewidth]{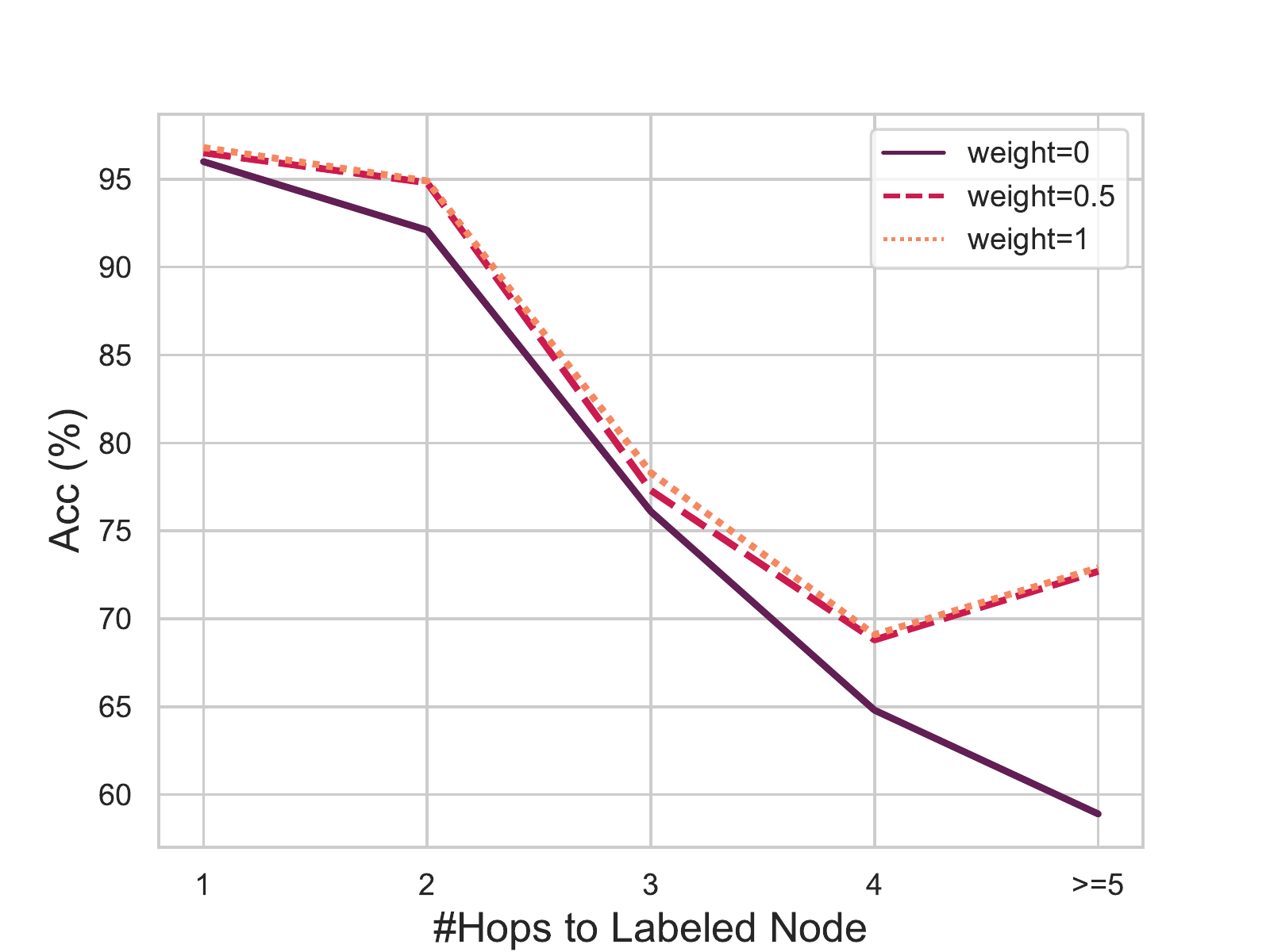}
  \subcaption{SAGE}
\end{minipage}%
\caption{The accuracy values of nodes sets with different hops to the labeled node. We display the result of all the four experiment models with different node pair loss weights ($0$/$0.5$/$1.0$). We can obviously find that for all models, adding node pair loss can help predict long-distance nodes better.}
\label{figure_long_distance}
\end{figure*}

\subsection{Comparison with State-of-the-art}
In this subsection, we display the comparison results between our \HGNN and state-of-the-art node classification methods in the setting of standard Planetoid~\citep{semi-supervised} dataset splitting for CORA/CiteSeer/PubMed datasets.  The base model used for \HGNN framework is GCN or GraphMix~\citep{model_graphmix}.
For baselines, we choose classical GCN and GAT and some recent advanced methods. The results are shown in Table~\ref{table_state-of-the-art}. We can observe that our \HGNN outperforms all the baselines and achieves state-of-the-art performance on CORA and CiteSeer. 
% Moreover, our \HGNN can also be applied to the recent advanced GraphMix ~\citep{model_graphmix} and further improve the node classification accuracy. This demonstrates the generalizability of \HGNN for both classical GNNs and recent advanced models.
Our approach also shows competitive results on PubMed.
The limited improvements on PubMed is because that there are only few labeled nodes (60) in the training set that provide inadequate supervision information for the node pair classification task (node pairs of the same category constitute a small part of all the node pairs, so there are few positive samples).

\subsection{Effects of Co-training}
To examine the effects of node pair classification,
we display 2D visualization of the node distribution belonging to different categories with varying loss weight $\lambda$ for node pair classification in Figure~\ref{figure_visual}. We train a 2-layer GNN and use T-SNE~\citep{tsne} to reduce hidden states to two dimensions.
From Figure~\ref{figure_visual}, we can observe that with the increase of $\lambda$, nodes of the same category are more concentrated and nodes of different categories are more separated, which makes GCN easier to set node classification boundaries with fewer prediction errors. Besides, with the increase of $\lambda$, the node classification accuracy can be generally improved. Thus, co-training with node pair classification indeed facilitates GCN training and enables to achieve higher accuracy.
We observe similar phenomena in other GNNs, but we only display the node distribution for GCN due to limited space.
\subsection{Results of the Long-distance Nodes}
Our \HGNN is designed to model long-distance node relations. To verify the effectiveness of \HGNN in modeling long-distance node relations, we split the test set into different subsets according to the topological distance between unlabeled nodes and labeled nodes, and present prediction results of each subset. Results of the four experiment GNNs are displayed in Figure~\ref{figure_long_distance}. We can observe that for all the GNNs, adding node pair loss can help predict long-distance nodes better, which verifies the effectiveness of \HGNN in modeling long-distance node relations.

\begin{table}[t]
\centering
% \small
\resizebox{.75\columnwidth}{!}{
\begin{tabular}{c|cccc}
\toprule
\textbf{Acc (\%)} & $\lambda$=0 & $\lambda$=0.5 & $\lambda$=1.0  & $\lambda$=2.0  \\ \hline
\textbf{10 Nodes}      & 78.5       & 78.8         & \textbf{79.3} & 78.5          \\ 
\textbf{20 Nodes}      & 78.6       & 80.8         & 82.5          & \textbf{82.8} \\ 
\textbf{30 Nodes}      & 80.6       & 84.2         & 84.6          & \textbf{84.8} \\  
\textbf{40 Nodes}      & 83.1       & 85.8         & \textbf{86.4} & 86.3          \\ 
\textbf{50 Nodes}      & 83.5       & 86.4         & 86.3          & \textbf{86.8} \\
\bottomrule

\end{tabular}}
\caption{Accuracy values (GCN model on the CORA dataset) of different node pair loss weight (x-axis) under different labeled node size per class (y-axis). We can find that our \HGNN can achieve consistent improvement with different training set size.}
\label{table_increase_node}
\end{table}

\begin{table}[t]
\centering
% \small
\resizebox{.95\columnwidth}{!}{
\begin{tabular}{c|ccccl}
\toprule
\textbf{Acc (\%)} & \textbf{Lead} & \textbf{Random} & \textbf{Close} & \textbf{Middle} & \textbf{Remote} \\ \hline
\textbf{100k}    & 85.9          & \textbf{86.3}   & 84.5           & 85.6            & 85.9            \\
\textbf{200k}    & 86.2          & 86.4            & 85.6           & 86.0            & \textbf{86.5}   \\
\textbf{300k}    & 86.2          & \textbf{86.7}   & 85.4           & 85.7            & 86.6            \\
\textbf{400k}    & 86.9          & \textbf{87.2}   & 85.8           & 85.7            & 86.5            \\
\textbf{500k}    & 86.4          & 86.0            & 85.6           & 85.4            & \textbf{86.9}  \\
\bottomrule
\end{tabular}}
\caption{Accuracy values (GCN model on the CORA dataset; 50 labeled nodes per category) of different sampling strategy (x-axis) under different sampling size (y-axis). ``Lead'' denotes sampling the front node pairs; ``Random'' means randomly sampling; ``Close'', ``Middle'', ``Remote'' denote sampling the node pairs with a higher probability for the node pairs with close, middle and remote topological distance, respectively. We can find that the ``Random'' strategy and the ``Remote'' strategy perform best among all strategies under different sampling numbers.}
\label{table_sample_strategy}
\end{table}

\subsection{Analysis of Node Pair Sampling}
Limited by the accessible nodes labels, our \HGNN  takes the node pairs from the training set for training. In this subsection, we conduct some extended analyses for the sampling process.
Firstly, in Table~\ref{table_increase_node}, we present the results of GCN with different
node pair loss weight $\lambda$ and labeled node size per class. We can find that our \HGNN can achieve consistent improvement with different training set size.
Secondly, in Table~\ref{table_sample_strategy}, we change sampling strategies under different total numbers of sampled node pairs. We set the labeled node size per class to be 50, so that we can design different sampling methods. The results show that the Random strategy and the Remote strategy perform best in different sampling numbers. We can also conclude that modelling the node pair relations for the remote nodes are more effective than modelling close nodes, because the long-distance relations are hardly learned by the original GNN models.

\section{Related Work}
The semi-supervised learning framework has been widely used on the node classification task~\citep{model_gcn,model_sage,model_cheb}. Recent works have proposed new advanced training methods. For example, \citet{jump_knowledge} propose jumping knowledge networks to utilize information from high-order neighbors.
\citep{model_graphmix} propose to apply the manifold mixup in GNN training to generate some virtual samples. \citet{drop_edge} propose to remove edges randomly at each epoch and acts as a data augmenter or a message passing reducer.

Modeling long-distance relations is vital for GNN training and applications. Recent studies~\citep{deepgcns,drop_edge,PairNorm} try to build deep GNN architectures or use the hyper-graph information~\citep{model_dna,model_hyper_graph,dynamic_hyper}. However, these solutions can hardly model the very long-distance node relations. Instead, our HighwayGraph framework can model the node relations regardless of their topological distance by simply relying on shallow GNN models.

Different GNN architectures~\citep{model_arma,model_feast,graph_seq} have been designed for graph-related tasks with different motivations. Most existing works directly use the original graph topology, while~\citet{chen_smoothing} prove that the performance of GNNs can be improved by graph topology optimization. Other works also refer to the dynamic graphs. \citet{dynamic-training} propose the EvolveGCN model by using the RNN model to update GCN. \citet{model_togcn} propose to train GCN and refine the graph topology at the same time. Some other works~\citep{GLCN,LEARNGraph} also try to learn and update the graph topology. Different from these works, our method optimizes graph with a clear target to build the information highway especially for remote nodes.

Self-training is a popular framework in the semi-supervised task as it can extend the supervised information based on prediction results. \citet{analysis_smoothing} propose to use predicted pseudo labels as the supervision information for the next training iteration. \citet{dynamic-training,GAM} follow and update this idea. Different from the method of using pseudo labels, our proposed HighwayGraph takes the advantage of predicted pseudo edges, which has proven significant performance improvements over multiple GNN models across benchmark graph datasets.

\section{Conclusion}
We propose to model long-distance node relations for graph structured data by simply relying on shallow GNNs. We provide two solutions: (1) Implicitly modelling node pairs by predicting their relations, and (2) Explicitly modelling node relations by adding edges.
Then we introduce a novel GNN training framework named \HGNN to combine these two solutions.
Extensive experimental results demonstrate that our proposed \HGNN consistently and significantly achieves improvements over multiple GNNs on three benchmark graph datasets with limited extra computational cost\footnote{Shown in Appendix.}, which verifies the effectiveness and generalization of our method. 
Besides, \HGNN enables to significantly improve prediction accuracy for unlabeled nodes that are far away from labeled nodes, which further justifies the two proposed solutions to modelling long-distance node relations.
In the future, we plan to find better methods in this direction.

\section{Acknowledgements}
This work was supported in part by a Tencent Research Grant. Xu Sun is the corresponding author of this paper.

\bibliography{acl2020}
\bibliographystyle{acl_natbib}

\appendix
\section{Additional Computational Cost of \HGNN}
Compared to the standard semi-supervised node classification, the module of graph topology optimization and the module of co-training in our \HGNN framework will cause inevitable additional computational cost. However, the additional computational cost is limited. 
\paragraph{(1) Graph Optimization Cost} In experiments, we observe  that a well-optimized graph topology is effective for consistently improving the performance of multiple GNNs on the same graph dataset, so our \HGNN can be easily transferred among different GNN models without extra computational cost for the graph topology optimization.

\paragraph{(2) Co-training Cost} Eq~9 is presented for a better understanding and can be optimized in practice. When training the node pair classification task, the supervision information is accessed from the training set of the node classification task, which is very small in the semi-supervised framework (20 nodes from each category as the training set; the whole training set usually has 60-200 nodes). Thus the size of training node pairs for the node pair classification task is also limited. When computing the training loss, we access the training set nodes first and then conduct matrix multiply in Eq~9, thus the computational complexity is $O(m \cdot m)$ instead of $O(n \cdot n)$ ($m$ denotes the training size and $n$ denotes the node size; usually $m \ll  n$). Besides, in the co-training module, the prediction of the node pair relation requires no extra trainable parameters and causes only a small increase in GPU memory occupancy.

\end{document}